# Mechanical Design Improvement of a Passive Device to Assist Eating in People Living with Movement Disorders

Michaël Dubé[1,2], Thierry Laliberté[1], Véronique Flamand[2,3], François Routhier[2,3], Alexandre Campeau-Lecours[1,2]

[1]Département de génie mécanique (Université Laval), [2]Centre interdisciplinaire de recherche en réadaptation et intégration sociale (CIRRIS), [3]Département de réadaptation (Université Laval)

## ABSTRACT

Many people living with neurological disorders, such as cerebral palsy, stroke, muscular dystrophy or dystonia experience upper limb impairments (muscle spasticity, loss of selective motor control, muscle weakness or tremors) and have difficulty to eat independently. The general goal of this project is to develop a new device to assist with eating, aimed at stabilizing the movement of people who have movement disorders. A first iteration of the device was validated with children living with cerebral palsy and showed promising results. This validation however pointed out important drawbacks. This paper presents an iteration of the design which includes a new mechanism reducing the required arm elevation, improving safety through a compliant utensil attachment, and improving damping and other static balancing factors.

## INTRODUCTION

Being able to eat without the assistance of a caregiver is an important aspect of independent daily living [1]. Unfortunately, many people living with conditions such as cerebral palsy, stroke, muscular dystrophy or dystonia experience movement disorders to the upper limbs (muscle spasticity, unselective motor control, muscle weakness or tremors) and have difficulty eating on their own. Numerous solutions have already been created to assist people living with such conditions. Liftware offers two intelligent handles to help people suffering from tremors or muscular stiffening, i.e. the Liftware Steady™ and Liftware Level™ handles [www.liftware.com], each available with the soup spoon, everyday spoon, fork and spork attachments. There are also mechanical devices to reduce the effect of spasms, e.g. the Neater Eater [www.neater.co.uk], the Action Arm [www.flaghouse.ca], the Friction Feeder [www.ncmedical.com] and the Nelson [www.focalmeditech.nl]. Some devices feed the users autonomously and require few actions, e.g. the iEAT Feeding Robot [www.assistive-innovations.com], the Winsford Feeder [2], and the OBI arm [www.meetobi.com]. Even if several solutions have been proposed or commercialized, the literature also points to a number of factors that limit the adoption of assistive technology (AT) devices in general: high cost, difficulties of operating devices, poor performance, and insufficient adaptation to the users' needs [3,4]. We also know, from scientific literature [1] and non-formal discussion with therapists, that many people living with movement disorders cannot eat independently and that the AT on the market are not suitable for their special needs and do not help them to eat by their own. This led to the creation of a first design for a mechanical AT that addresses two types of motor disorders: a) contractures due to spasticity or joint deformities which prevent the user from holding the utensil parallel to the ground, and b) abrupt movements such as spasms, ataxia or dystonia. The device was designed to stabilize the user's motion and to enable independent eating. Once developed, the device was tested in a trial with potential users [1,5]. Occupational therapists supervised the trial and noted different improvements that could be brought on to the prototype. The main comments were that a) bringing the utensil to the mouth with the device required too much arm elevation, thus raising safety issues relative to the utensil attachment (i.e. if the utensil is rigidly attached to the mechanism, it could hurt the user if he/she makes an involuntary movement while the utensil is in the mouth) b) the motion damping presents a dead-zone (there was no damping for small movements), and c) the static balancing of the mechanism weight was not optimal. The work described in the current paper addresses these issues through the design of mechanical improvements. The first section hosts the description of the previous design, followed by the objectives of the project, and a summary depiction of the device. Then, each improvement to the first prototype is detailed: the handle attachment, the compliant spoon design, the static balancing, and the new dampers. Finally, the work is shortly discussed and concluded.

## PREVIOUS DESIGN

This section presents the general mechanical design of both the first prototype and the new one. Fig. 1 presents, in order of increasing complexity (Fig. 1a, b, and c, respectively), three variations of the mechanism, all of which allow the same three degrees of freedom (DoF). Fig. 1a shows a simple system with three pivots (J1, J2 and J3), which is known as an RRR (three rotary joints) mechanism. The parallelogram added in Fig. 1b is used to damp L2-bar's rotation around J3. L1-bar is damped using the J2 pivot. Fig. 1c shows the complete assembly with the two other parallelograms used to maintain the orientation of the spoon [1]. This design was also used as a basis for the development of a writing assistive device [6]. Fig. 2a depicts the first iteration of the prototype while Fig.2b displays the prototype with the modifications that will be described in the following sections.

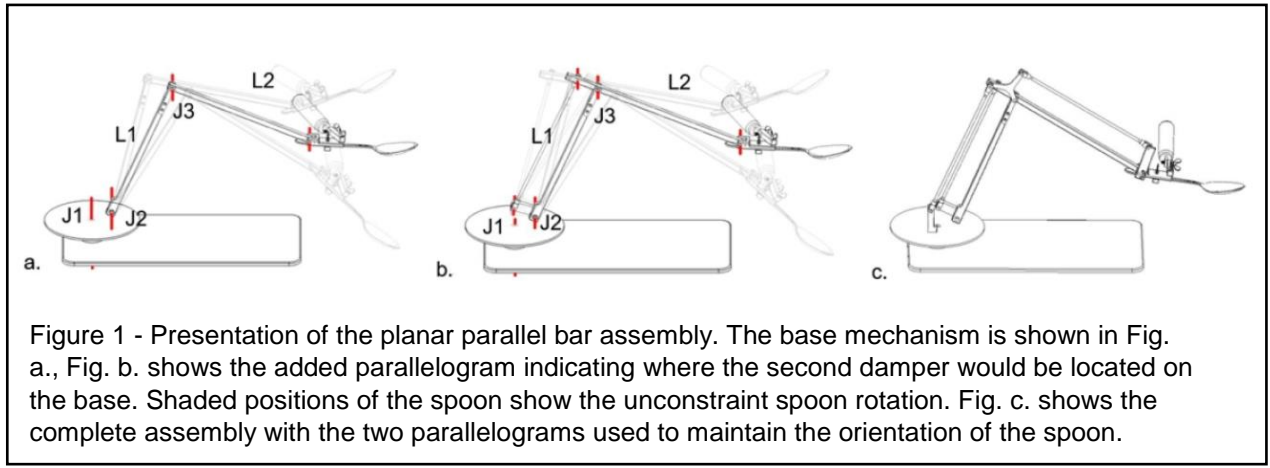

Figure 1 - Presentation of the planar parallel bar assembly. The base mechanism is shown in Fig. a., Fig. b. shows the added parallelogram indicating where the second damper would be located on the base. Shaded positions of the spoon show the unconstraint spoon rotation. Fig. c. shows the complete assembly with the two parallelograms used to maintain the orientation of the spoon.

## OBJECTIVES

The main objective of the project is to design an improved version of an assistive eating device prototype aimed at stabilizing the motion of people living with movement disorders. More specifically, based on a previous iteration, four features are addressed in this work: 1) reducing the required shoulder elevation, 2) making the spoon compliant to avoid injuring the user if he/she has a head spasm while eating, 3) statically balancing the mechanism and 4) exploring a new kind of damper in the prototype.

## SUMMARY DESCRIPTION

The proposed mechanism, which is designed to be mounted on a table, is shown in Fig. 2b. The mechanism has three DoFs (J1, J2, J3 in Fig.1a). A spoon is attached at the end of the mechanism. The user operates the device by grasping and moving a handle. The orientation and the height of the handle can be adjusted to the user's preference, inside a predetermined range (detailed in the next section). The device allows moving the spoon in every direction inside the working area and, as a result of the mechanism design, maintains the spoon and the handle in a constant orientation. The spoon orientation can be changed depending on the food that the user is eating. Mechanical inertia and dampers allow stabilization of the user's motion. The device thus assists the user in two different manners. First, by holding the spoon in the same orientation, it facilitates a task that is difficult or impossible for some people because of spasticity or upper limb impairments. Second, the added inertia and damping stabilize uncoordinated movements (i.e. spasms). Although the mechanism is shown here with a spoon attached to it, a fork can also be used. A scooper plate with a suction cup base is attached to the mechanism.

This prototype has already seen a first design, which will be presented, followed by the four major improvements brought on to the new version of the prototype: 1) a modified handle attachment to reduce arm elevation requirements, 2) a compliant spoon attachment, 3) the static balancing of the mechanism, 4) a new type of damper.

## HANDLE ATTACHMENT DESIGN

The main difference between the previous design (Fig.2a) and the novel design (Fig.2b) is a new handle mechanism which allows the user to reach his/her mouth with less arm elevation amplitude. This was an important suggestion from the occupational therapists in the previous design trials that will help users who do not have the capacity (or have difficulty) to raise their hand up to their mouth (majority of users tested). By fixing the handle within the parallelogram (Fig.2b) instead of fixing it at the end of it (Fig.2a), the required upward handle movement amplitude to generate the same spoon vertical movement is much reduced. By comparison, with the previous design, raising the utensil by 33 cm (average mouth height from the table) required the same handle upward motion. With the proposed design, the required upward motion of the handle is 24 cm. The handle's bracket is attached to the parallelogram farther from the utensil and thus makes the movement reduction possible.

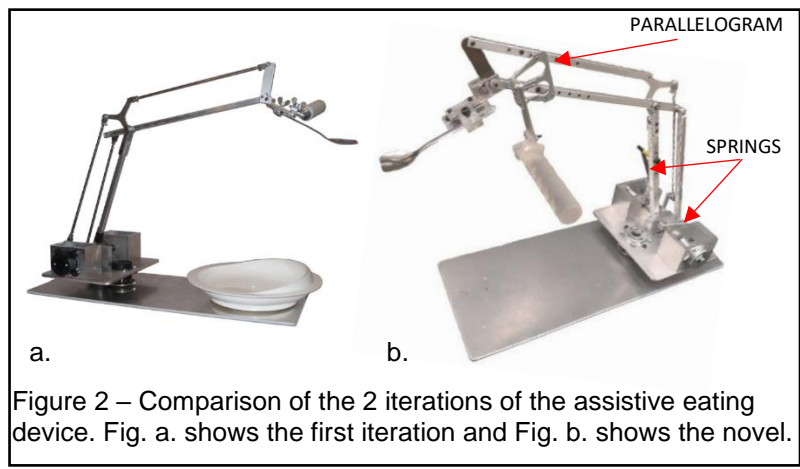

Figure 2 – Comparison of the 2 iterations of the assistive eating device. Fig. a. shows the first iteration and Fig. b. shows the novel.

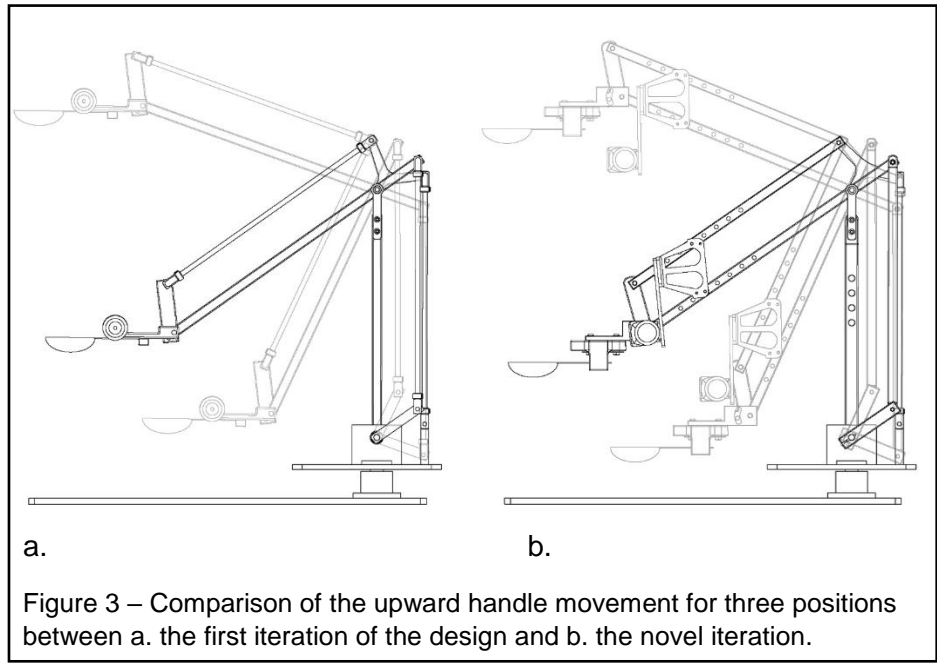

a.  b.

Figure 3 – Comparison of the upward handle movement for three positions between a. the first iteration of the design and b. the novel iteration.

The bracket positions the handle higher than the utensil when the utensil is in the plate, and it positions the handle lower than the utensil when the utensil is at the user's mouth level. The bracket is L-shaped to position the handle closer to the user. Another bracket is also positioned between the L-shape bracket and the handle to lower the handle and make it easier for the user to manipulate it. The bracket can be assembled for both right-handed and left-handed users. The orientation of the handle can also be set to five different angles to meet the user's needs. For the user's comfort, the handle is free to rotate around its own axis but can also be locked with a small socket head screw.

**COMPLIANT SPOON ATTACHMENT DESIGN**

During the tests, occupational therapists raised safety issues relative to the utensil attachment (i.e. if the utensil is rigidly attached to the mechanism, it could hurt the user if he/she makes an involuntary movement while the utensil is in the mouth). A compliant utensil attachment was thus designed. The compliance gives a flexibility so that it does not hurt the person in such a case. The utensil attachment's compliant DoFs are presented in Fig.4. Parts #1 and #2 in Fig.4 are fastened together with two loose socket head screws enveloped with rubber. The rubber damps the shock if the user has a spasm and hits himself/herself. It is also used to re-center the spoon. In this design, it is important that the utensil moves if it is hit by the user. However, for intuitiveness purposes, it must also move back to the center position afterwards (otherwise, the user will always be chasing the utensil, which could be in a different orientation). To this end, a compliant utensil attachment was designed. It consists of flexible rubber cushions that can deform and absorb shock if there is a contact, and that will also move the utensil back to the center position afterwards.

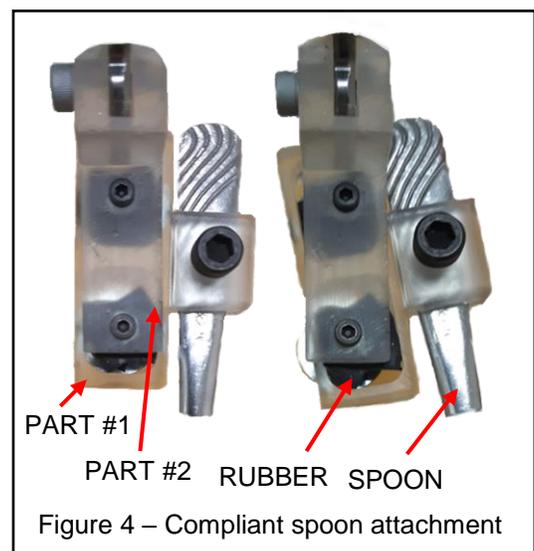

PART #1
PART #2  RUBBER  SPOON
Figure 4 – Compliant spoon attachment

**STATIC BALANCING**

The previous version of the prototype used 2 torsion springs to balance the mechanism but it did not prove efficient enough. To help the user use the prototype, the mechanism is balanced by 2 linear springs, their placement is shown on Fig.2b. With the springs, the user doesn't have to lift the weight of the mechanism, it is therefore easier when a user tries to take food from the plate to his/her mouth. This improvement could be useful for users who do not have much arm or hand strength (e.g. the elderly).

**INCLUSION OF NEW DAMPERS**

Another point that was revealed during the trials with the previous design was that the dampers were not good to stabilize the beginning of the motion although they were correct for the rest of the motion. Several commercial dampers were bought from ACE Controls Inc. and were tested (FRT-F2-203, FRT-F2-403, FDT-47, FDT-57) and the one that meets the needs of this application the most is the FDT-47 damper [7]. Since the design of the chosen damper is different from the previous one, adjustments had to be made to ensure a right fit on the mechanism. The two kinds of dampers can be seen on Fig.2, where the old ones are black and the new ones are metallic. The new dampers improved the fluidity of the motion assistance.

**DISCUSSION AND CONCLUSION**

In this paper, a novel design of a 3-DoF assistive eating device was presented. The device is used to support people living with movement disorders. The objectives were to develop an improved version of the first design iteration of the prototype based on a trial with potential users. The modifications made on the mechanism include the redesign of the handle and spoon attachment, the improvement of the static balancing, and a new kind of damper. Future work will consist in evaluating the novel prototype with potential users in order to assess the efficiency of the improvements in the eating process compared to the last version.

**ACKNOWLEDGEMENTS**

This work is supported by Dr. Alexandre Campeau-Lecours's research funds at Centre interdisciplinaire de recherche en réadaptation et intégration sociale (CIRRIS) and Université Laval (Québec City, Canada).

**Alternative text**

Figure 1: Figure 1 presents three variations of the mechanism, each one has a different level of complexity. The first variation (Fig. 1a.) shows a round plate linked to the base of the mechanism by the joint J1. A bar (L1) is attached to the plate by the joint J2. Another bar (L2), is linked at the end of the L1-bar with the joint J3. The spoon and the handle are attached at the end of the L2-bar. The second variation (Fig. 1b.) presents the same element that are in Fig. 1a. but bars are added to form a parallelogram with the L1-bar. The third variation (Fig. 1c.) presents a higher level of complexity than Fig. 1b., other parts are added to form a parallelogram with L2-bar. On Fig. 1a and 1b, shaded positions of the mechanism show the unconstraint spoon rotation.

Figure 2: Figure 2a presents the old version of the mechanism and Figure 2b shows the novel. On Fig. 2b. arrows indicate the location of the springs, which are attached from the base of the mechanism to the bars linked directly to the base. The differences between the old and the new version are the following; the spoon attachment of the new version is flexible (and made with plastic instead of aluminium), unlike the old version, the handle attachment is now fixed within the parallelogram instead of at the end of it, the dampers are different and springs are installed in the new version.

Figure 3: Figure 3a presents a side view of three positions of the first version of the mechanism. Figure 3b presents the same positions with the new version. With the old version, the handle has to move more than with the new version to create the same spoon movement.

Figure 4: Presents the compliant spoon attachment design in two different positions from the same point of view. It shows the location of the spoon and the rubber.